\documentclass[10pt,twocolumn,letterpaper]{article}

\usepackage{cvpr}              

\usepackage{graphicx}
\usepackage{amsmath}
\usepackage{amssymb}
\usepackage{booktabs}
\usepackage{multirow}

\usepackage[pagebackref,breaklinks,colorlinks]{hyperref}

\usepackage[capitalize]{cleveref}
\crefname{section}{Sec.}{Secs.}
\Crefname{section}{Section}{Sections}
\Crefname{table}{Table}{Tables}
\crefname{table}{Tab.}{Tabs.}

\def\cvprPaperID{7354} 
\def\confName{CVPR}
\def\confYear{2022}

\begin{document}

\def\ie{\emph{i.e.}}
\def\eg{\emph{e.g.}}
\def\wrt{\emph{w.r.t.}}

\definecolor{purp}{rgb}{0.65, 0.16, 0.65}
\definecolor{bblue}{rgb}{0.2, 0.2, 0.6}
\definecolor{green}{rgb}{0, 0.5, 0}
\definecolor{red}{rgb}{0.5, 0, 0}
\newcommand{\ahyun}[1]{{\color{bblue}{#1}}}
\newcommand{\mcho}[1]{{\color{magenta}{#1}}}

\newcommand{\Fig}[1]{Fig.~\ref{fig:#1}}
\newcommand{\Sec}[1]{Sec.~\ref{sec:#1}}
\newcommand{\Eq}[1]{Eq.~(\ref{eq:#1})}
\newcommand{\Tbl}[1]{Tab.~\ref{tab:#1}}
\newcommand{\Alg}[1]{Algorithm \ref{algo:#1}}

\newcommand{\Model}{EquiSym} 
\newcommand{\Dataset}{DENDI} 

\title{Reflection and Rotation Symmetry Detection via Equivariant Learning}

\author{Ahyun Seo \quad Byungjin Kim \quad Suha Kwak \quad Minsu Cho\vspace{0.15cm}\\
Pohang University of Science and Technology (POSTECH), South Korea\\
{\small \url{http://cvlab.postech.ac.kr/research/EquiSym}}
}

\maketitle

\begin{abstract}
The inherent challenge of detecting symmetries stems from arbitrary orientations of symmetry patterns; a reflection symmetry mirrors itself against an axis with a specific orientation while a rotation symmetry matches its rotated copy with a specific orientation. 
Discovering such symmetry patterns from an image thus 
benefits from an equivariant feature representation, which varies consistently with  reflection and rotation of the image.
In this work, we introduce a group-equivariant convolutional network for symmetry detection, dubbed EquiSym, which leverages equivariant feature maps with respect to a dihedral group of reflection and rotation.
The proposed network is built end-to-end with dihedrally-equivariant layers and trained to output a spatial map for reflection axes or rotation centers.
We also present a new dataset, \textit{DENse and DIverse symmetry} (DENDI), which mitigates limitations of existing benchmarks for reflection and rotation symmetry detection.
Experiments show that our method achieves the state of the arts in symmetry detection on LDRS and DENDI datasets.
\end{abstract}

\section{Introduction}
\label{sec:intro}
From molecules to galaxies, from nature to man-made environments, symmetry is everywhere.
Comprehensive perception and exploitation of real-world symmetry are the instinctive abilities of humans and animals that have the potential to take intelligent systems to the next level.
The focus of this paper is on the two most primitive symmetries, reflection and rotation symmetries.
The goal of reflection and rotation symmetry detection is to find a reflection axis and a rotation center that remain invariant under reflection and rotation, respectively.
Despite decades of efforts~\cite{weyl1952symmetry, liu2010computational}, symmetry detection methods have been limited to the well-defined symmetry patterns, and the remedy for real-world symmetry is still yet to be thoroughly explored.
\begin{figure}[t]
    \centering
    \includegraphics[width=0.47\textwidth]{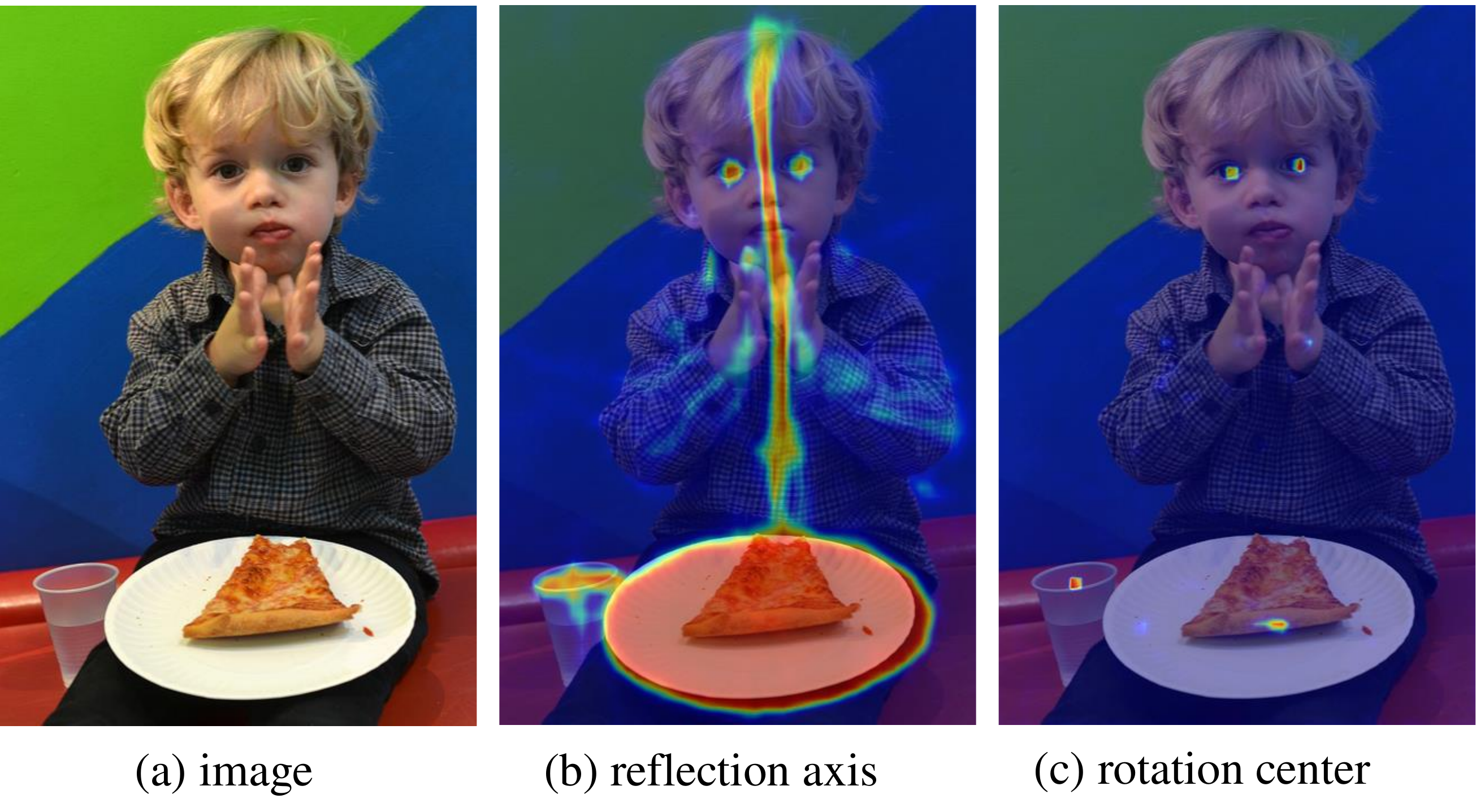}
    \caption{Symmetry detection examples of our method EquiSym. 
    (a) an input image, (b) a score map of reflection symmetry axes, and (c) that of rotation symmetry centers. Best viewed in color.}
    \label{fig:teaser}
    \vspace{-2mm}
\end{figure}

The simplicity of mathematical concepts of symmetry encouraged early approaches to find keypoint pairs that satisfy pre-defined constraints for symmetry~\cite{loy2006detecting, prasad2005detecting, shen2001robust, wang2015reflection, atadjanov2016reflection, cho2009bilateral}, which leverage hand-crafted local feature descriptors to detect sparse symmetry patterns. 
Recently, convolutional neural networks (CNNs) have been successfully applied to detect reflection symmetry and have surpassed the previous methods by learning score map regression~\cite{funk2017beyond} or symmetric matching~\cite{seoshim2021pmcnet} from data.


The primary challenge in detecting symmetry patterns lies in the fact that 
a symmetry manifests itself with an arbitrary orientation and perceiving the pattern requires an analysis based on the orientation; a reflection symmetry mirrors itself against an axis with a specific orientation and a rotation symmetry matches its rotated copy with a specific orientation. 
Most methods for symmetry detection thus involve searching over the space of candidate orientations of symmetry patterns and also developing a robust representation that is either {\em invariant} or {\em equivariant} with respect to rotation and reflection. 
The early approaches leverage an equivariant representation by extracting oriented keypoints and performing orientation normalization~\cite{loy2006detecting, prasad2005detecting, shen2001robust, wang2015reflection, atadjanov2016reflection}. 
While this technique has proven effective for shallow gradient-based features, it cannot be applied to deep feature maps from standard neural networks, where rotation and reflection induce unpredictable variations in representation. 

To address the challenge, we propose to learn a group-equivariant convolutional neural network for reflection and rotation symmetry detection, dubbed {\em {\Model}}.
Recently, there has been active research on equivariant networks to incorporate equivariance explicitly for robust and sample-efficient representation learning~\cite{cohen2016group, hoogeboom2018hexaconv, cohen2016steerable, worrall2017harmonic, sosnovik2019scale, e2cnn}. 
Unlike standard neural networks, they induce predictable and structure-preserving representation with respect to the geometric transformations, \eg, rotation or reflection, which is eminently suitable for symmetry detection.
To detect consistent symmetry patterns over different orientations, we build a dihedrally-equivariant convolutional network~\cite{e2cnn}, which is designed to be end-to-end equivariant to a group of reflection and rotation.
The network effectively learns to output a score map of reflection axes for reflection symmetry or that of rotation centers for rotation symmetry. 

We also present a new dataset, \textit{DENse and DIverse symmetry} ({\Dataset}), for reflection and rotation symmetry detection.  
{\Dataset} contains real-world images with accurate and clean annotations for reflection and rotation symmetries and mitigates limitations of existing benchmarks~\cite{liu2013symmetry, ConvSymm2016, seoshim2021pmcnet, funk2017beyond, funk20172017}. 
First, the reflection symmetry axes are diverse in scale and orientation,
while previous datasets mostly focus on the dominant axes of the vertical or horizontal ones.
Second, the rotation centers are annotated to the objects in polygon and ellipse shape, 
not limited to the circular objects.
Third, the number of the rotation folds for each rotation center is annotated, which is the first in a large-scale dataset. 
Finally, the number of images is 1.7x and 2.0x larger than the second-largest reflection and rotation symmetry detection datasets, respectively.

The contribution of our work can be summarized as:
\vspace{-1mm}
\begin{itemize}
    \item We propose a novel group-equivariant symmetry detection network, {\Model}, which outputs group-equivariant score maps for reflection axes or rotation centers via end-to-end reflection- and rotation-equivariant feature maps. 
	\vspace{-1mm}
	\item We present a new dataset, DENse and DIverse symmetry dataset ({\Dataset}), containing images of reflection and rotation symmetries annotated in a broader range of typical real-world objects.
	\vspace{-1mm}
	\item We show the outstanding performance of {\Model} in reflection symmetry detection on SDRW~\cite{liu2013symmetry}, LDRS~\cite{seoshim2021pmcnet}, and  {\Dataset}, and in rotation symmetry detection on {\Dataset}.
\end{itemize}

\section{Related Work}
\subsection{Equivariant deep learning}
Equivariance is a desirable inductive bias that improves generalization and sampling efficiency.
The conventional convolution is equivariant to translations but not to other transformations such as rotations and reflections.
Group equivariant CNNs~\cite{cohen2016group, hoogeboom2018hexaconv} use group convolution to learn equivariant representations for symmetry groups.
Marcos~\etal~\cite{marcos2017rotation} generate and propagate vector fields that maintain the maximum response along with the corresponding direction throughout the network.
Worrall~\etal~\cite{worrall2017harmonic} exploit circular harmonics to obtain rotational equivariance in a continuous domain.
Cohen~\etal~\cite{cohen2016steerable} combine fixed base filters linearly, resulting in steerable filters with no interpolation artifacts.
Equivariant CNNs on homogeneous spaces~\cite{weiler20183d,cohen2018intertwiners,cohen2018general,cohen2019gauge} are also proposed.
The aforementioned methods consider equivariance to specific transformations until Weiler~\etal~\cite{e2cnn} provide a general solution of kernel space constraint for arbitrary group representations of the Euclidean group E(2).
From the perspective of an application, Han~\etal~\cite{han2021redet} and Gupta~\etal~\cite{gupta2021rotation} extract rotation-equivariant feature maps for oriented object detection and visual tracking, respectively. 
We leverage E(2)-equivariant CNNs~\cite{e2cnn} as a building block of our network to perceive consistent symmetry patterns across multiple orientations.

\subsection{Symmetry detection}

Symmetry detection deals with different kinds of symmetric patterns such as
reflection axis~\cite{loy2006detecting, atadjanov2016reflection, prasad2005detecting, wang2014unified, wang2015reflection, funk2017beyond, FUKUSHIMA20061827, seoshim2021pmcnet, tsogkas2012learning, gnutti2021combining, keller2006signal}, 
rotation center~\cite{loy2006detecting, funk2017beyond, wang2014unified, lee2008rotation, prasad2005detecting, lee2009skewed, keller2006signal}, and
translation lattice~\cite{liu2004computational, lin1997extracting, wang2014unified, liu2004near, zhao2011translation, hays2006discovering, park2009deformed}.

\paragraph{Sparse prediction.}
Rotation and translation symmetries are often formulated as periodic signals and detected by autocorrelation~\cite{lin1997extracting, liu2004computational} in the spatial domain, spectral density~\cite{lee2008rotation, lee2009skewed} and angular correlation~\cite{keller2006signal} in the frequency domain. 
Meanwhile, there is a consistent need for an affine-invariant or equivariant feature descriptor in detecting symmetries, as matching the local descriptors is the most common solution.
Loy and Eklundh~\cite{loy2006detecting} use SIFT~\cite{lowe2004distinctive} descriptors and normalize each descriptor by its dominant orientation.
Cho and Lee~\cite{cho2009bilateral} also use SIFT~\cite{lowe2004distinctive} to match feature pairs and detect symmetry by discovering the clusters of the nearby matches.
Contour and edge features~\cite{shen2001robust, atadjanov2016reflection, prasad2005detecting, wang2014unified, wang2015reflection} are also useful for determining the boundary of the symmetric object. 
Lee and Liu~\cite{lee2009skewed} propose to solve affine-skewed rotation symmetry group detection by rectifying the skewed patterns. 
In this paper, we tackle the task by data-driven approach with our proposed dataset.
Also, we use the dihedral group to interpret the symmetries as in many symmetry detection literatures~\cite{lee2008rotation, lee2009skewed}.

\begin{figure*}[t!]
    \centering
    \includegraphics[width=\textwidth]{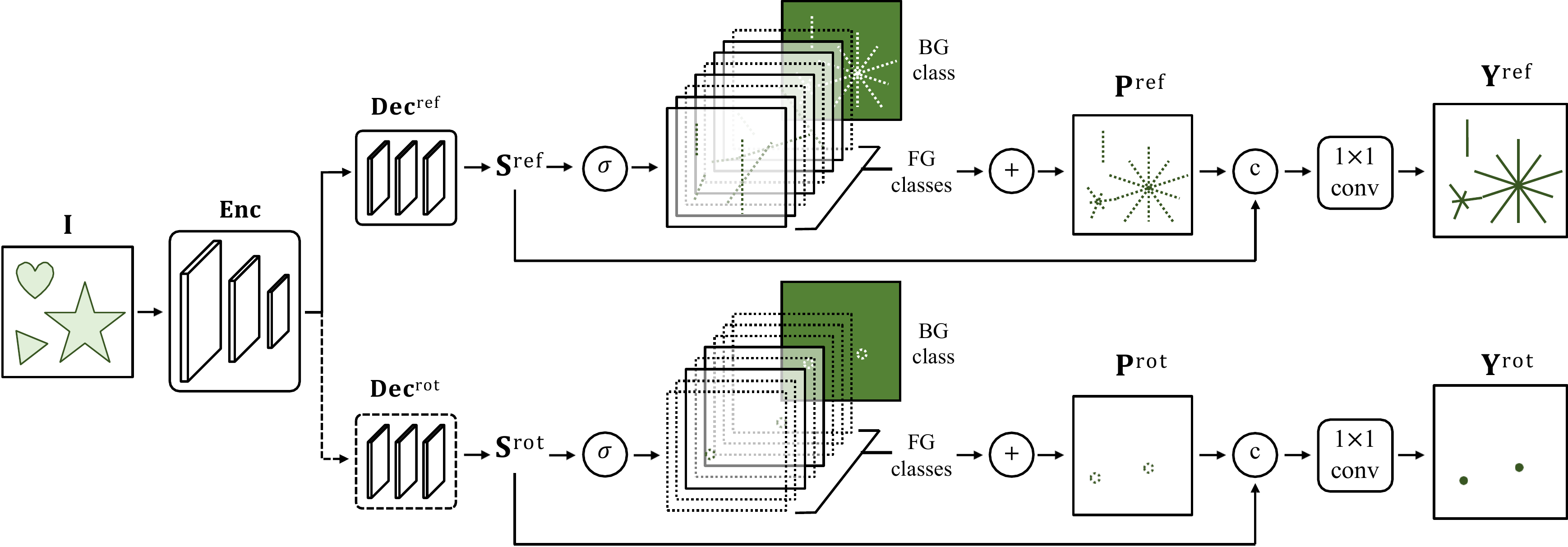}
    \caption{Illustration of the proposed symmetry detection network, {\Model}. 
    After an input image $\mathbf{I}$ passed a group-equivariant encoder $\mathrm{Enc}$, group-equivariant decoders $\mathrm{Dec}^\mathrm{ref}$ and $\mathrm{Dec}^\mathrm{rot}$ predict intermediate predictions $\mathbf{S}^\mathrm{ref}$ and $\mathbf{S}^\mathrm{rot}$ for rotation and reflection, respectively. 
    Auxillary tasks for the rotation and reflection symmetry are the order(N) of the rotation fold and the orientation of the reflection axis. The foreground logits are pooled to $\mathbf{P}^\mathrm{ref}$ and $\mathbf{P}^\mathrm{rot}$ and stacked with the scores $\mathbf{S}^\mathrm{ref}$ and $\mathbf{S}^\mathrm{rot}$, respectively.
    The final score $\mathbf{Y}^\mathrm{ref}$ and $\mathbf{Y}^\mathrm{rot}$ 
    for the rotation center and the reflection axis are predicted 
    using a group-equivariant $1\times1$ convolution.
    For details, see \Sec{method}.
    }
    \vspace{-2mm}
    \label{fig:infer}
\end{figure*}

\paragraph{Dense prediction.}
Recently proposed methods~\cite{FUKUSHIMA20061827, tsogkas2012learning, funk2017beyond, seoshim2021pmcnet} predict pixel-wise symmetry scores. 
Fukushima and Kikuchi~\cite{FUKUSHIMA20061827} build a neural network to extract edges from images and detect reflection symmetry.
Tsogkas \etal~\cite{tsogkas2012learning} construct a bag of features using histogram, color, and texture for each pixel and adopt multiple instance learning when training the model.
Gnutti~\etal~\cite{gnutti2021combining} take two stages of computing the symmetry score for each pixel using patch-wise correlation and validating the obtained candidate axes using gradient direction and magnitude.
Funk and Liu~\cite{funk2017beyond} are the first to adopt deep CNNs 
for detecting reflection and rotation symmetries. 
Seo~\etal~\cite{seoshim2021pmcnet} propose a polar self-similarity descriptor
for better rotation and reflection invariance. A specially designed Polar Matching Convolution (PMC) performs region-wise feature matching to compute the symmetry score, but the model relies heavily on the CNNs.
To discover consistent symmetry patterns {\wrt} geometric transformations of rotation and reflection, we deploy group equivariant neural networks in our symmetry detection model.


\section{Proposed Method}
\label{sec:method}
The symmetry patterns appearing in an image are invariant to the 2D rigid transformations of the image. 
The detected symmetry patterns of a transformed input image should be the same with the transformed detection results using the original input image.
In other words, the reflection and rotation equivariance is crucial for a symmetry detection model. 
To this end, we propose a unified framework for detecting the reflection and rotation symmetries via equivariant learning, {\Model}.
The overall pipeline is briefly illustrated in ~\Fig{infer}.
Given an input image, a shared encoder $\mathrm{Enc}$ and decoders $\mathrm{Dec}^\mathrm{ref}$ and $\mathrm{Dec}^\mathrm{rot}$ are applied for detecting reflection and rotation symmetries, respectively.
We also perform auxiliary pixel-wise classification tasks, one for the orientation of the reflection axis and the other for the order of the rotation symmetry.
The intermediate logits and the corresponding probabilities of the subtasks are denoted by ${\mathbf{S}}$ and ${\mathbf{P}}$, respectively.
The logits ${\mathbf{S}}$ are integrated with the sum of the foreground probabilities ${\mathbf{P}}$ to compute the final score map ${\mathbf{Y}}$ using a $1\times1$ group-equivariant convolution layer.
The following sections cover the preliminaries and the proposed symmetry detection network.

\subsection{Preliminaries}
\paragraph{Group and equivariance.}
A group $(G, \cdot)$ is a set $G$ with a binary operation $\cdot$, where its elements are closed under the operation.
A group has unique identity element and inverse element and also satisfies the associativity.
Equivariance of a map $f: X \xrightarrow{} Y$ is formalized using a group $G$ and two $G$-sets $X$ and $Y$, where $G$-set is a mathematical object consisting of a set $S$ and a group action of $G$ on $S$.
A map $f$ is said to be equivariant iff
\begin{align}
    f(g \cdot x) = g \cdot f(x),
\end{align}
for all $x \in X$ and all $g \in G$.
In 2D image domain, we focus on an euclidean group E(2), which is a group of plane $\mathbb{R}^2$ isometries of translations, rotations, and reflections. 


\paragraph{E(2)-steerable feature field.}
The affine transformations of the 2D or 3D coordinates are easily done by matrix multiplications.
Unlike low-dimensional feature vectors, the reflection or rotation transformation of the high-dimensional feature vectors are non-trivial.
The first step to build a \textit{steerable} convolution is to define a \textit{steerable} feature field
$f : \mathbb{R}^2 \xrightarrow{} \mathbb{R}^c$ that maps a feature vector $f(x) \in \mathbb{R}^c$ with each point x of a base plane.
Given a group $G$, a group representation $\rho : G \xrightarrow{} \mathrm{GL}(\mathbb{R}^c)$ specifies the transformation law for shuffling the $c$ channels of each feature vector, where a general linear group $\mathrm{GL}$ is the set of $c \times c$ invertible matrices.
Thus, applying transformation on a feature map not only moves the target vectors to the new positions but also shuffles each vector via $\rho(g)$ where $g \in G$. 
The group representations of E(2) group are presented in ~\cite{e2cnn}. 


\paragraph{E(2)-equivariant steerable convolution.}
To preserve the transformation law of the steerable feature spaces in CNNs,
equivariance under the group actions is required for each network layer.
Convolutions with the restricted \textit{G-steerable kernels}~\cite{e2cnn} provide an equivariant linear mapping between the steerable feature spaces. 
The input and output of the \textit{G-steerable} layers are the feature fields with their group representations $\rho_{\mathrm{in}}(g) \in \mathbb{R}^{c_{\mathrm{in}} \times c_{\mathrm{in}}}$ and $\rho_{\mathrm{out}}(g) \in \mathbb{R}^{c_{\mathrm{out}} \times c_{\mathrm{out}}}$,
where a group element $g$ is specified.
A kernel $k : \mathbb{R}^2 \xrightarrow{} \mathbb{R}^{c_{\mathrm{out}} \times c_{\mathrm{in}}}$ that transforms under 
$\rho_{\mathrm{in}}$ and $\rho_{\mathrm{out}}$ becomes \textit{G-steerable} when satisfying a kernel constraint of
\begin{align}
    k(gx)    &= \rho_{\mathrm{out}}(g) k(x) \rho_{\mathrm{in}}(g^{-1}),
\end{align}
for every $g \in G$ given $x \in \mathbb{R}^2$.
E(2)-equivariant CNNs solve this constraint to get a basis of the steerable kernels and comput the convolutional weights, which results in the smaller learnable parameters compared to the CNNs.

\subsection{Symmetry detection network}
\paragraph{Reflection and rotation equivariant modules.}
Since we aim to establish both reflection and rotation symmetry, we employ an E(2)-equivariant CNNs of dihedral group $\mathrm{D}_N$, which contains $N$ discrete rotations by angles multiples of $\frac{2 \pi}{N}$ and reflections.
The encoder $\mathrm{Enc}$ consists of an E(2)-equivariant~\cite{han2021redet, e2cnn} ResNet~\cite{he2016deep} and an Atrous Spatial Pyramid Pooling(ASPP)~\cite{DBLP:journals/corr/ChenPSA17} module. 
The decoder $\mathrm{Dec}$ is a 3-layer convolution module. 
The encoder and decoder designs follow~\cite{seoshim2021pmcnet} except that all convolution layers are substituted by the E(2)-equivariant convolution layers.
During the forward computation, the feature fields are transformed into the predefined fields of the group $\mathrm{D}_N$.
For the predictions ${\mathbf{S}}^\mathrm{ref}$ and ${\mathbf{S}}^\mathrm{rot}$, the feature fields of the decoders $\mathrm{Dec}^\mathrm{ref}$ and $\mathrm{Dec}^\mathrm{rot}$ are transformed back to the scalar fields.

\paragraph{Auxiliary classification.}
Instead of a direct regression of the symmetry score maps~\cite{funk2017beyond, seoshim2021pmcnet}, we perform relevant subtasks that can lead to the final prediction.
The proposed auxiliary tasks are the pixel-wise classification of the orientation (angle) of the reflection axis and the number of rotation folds. 
For simplicity, we assign the orientation into $N^\mathrm{ref}$ bins dividing 180 degrees.
The ground-truth orientation $\mathbf{S}^\mathrm{ref}_\mathrm{gt}$ is then quantized to be one-hot.
The auxilary rotation label $\mathbf{S}^\mathrm{rot}_\mathrm{gt}$ is the order of the rotational symmetry, which is annotated with a positive integer in the case of the discrete rotational symmetry.
We allocate '0' to the continuous group since its ground-truth order is infinite.
The size of the unique set of the rotation orders in the dataset is denoted as $N^\mathrm{rot}$.
Meanwhile, we add background classes for the pixels that are neither the axis nor the center.
Therefore, the classifiers predict the scores of channel of $N^\mathrm{ref}+1$ and $N^\mathrm{rot}+1$.
The classification logit ${\mathbf{S}} \in \mathbb{R}^{H \times W \times (N + 1)}$ is obtained by
\begin{align}
    {\mathbf{S}} &= \mathrm{Dec} (\mathrm{Enc} (\mathbf{I})).
\end{align}
The encoder $\mathrm{Enc}$ is shared while the decoders $\mathrm{Dec}^\mathrm{ref}$ and $\mathrm{Dec}^\mathrm{rot}$ are task-specific. 
Note that we set the group orientation $N$ as the same as $N^\mathrm{ref}$ to further exploit the equivariance of the equivariant networks.

\paragraph{Symmetry detection.}
The predicted score maps of the symmetry axes of the corresponding orientation are present in the $N^\mathrm{ref}$ foreground channels of the estimated orientation ${\mathbf{S}}^\mathrm{ref}$, whereas the background channel contains the background pixels.
Similar to reflection, the $N^\mathrm{rot}$ foreground channels are the score maps of the rotation centers with that number of folds. 
We aggregate the sum of the foreground logits $\mathbf{P}\in \mathbb{R}^{H \times W \times 1}$ and the intermediate prediction ${\mathbf{S}} \in \mathbb{R}^{H \times W \times (N + 1)}$ to compute the final prediction ${\mathbf{Y}}\in \mathbb{R}^{H \times W}$ as
\begin{align}
  {\mathbf{P}}_{h, w} &= \sum^{N}_{k=1} \frac{\exp{({\mathbf{S}}_{h, w, k})}} 
                    {\sum_{c} \exp{({\mathbf{S}}_{h, w, c})}},  \\
  {\mathbf{Y}} &= \mathrm{conv}_\mathrm{G}([\mathbf{P} || \mathbf{S}]).
\end{align}
Note that $||$ denotes the concatenation operation along the final channel dimension.

\subsection{Training objective}
To train {\Model}, we optimize a combination of two loss terms for localization and classification.
Following~\cite{seoshim2021pmcnet}, we adopt the focal loss~\cite{lin2017focal} as the localization loss $\mathcal{L}_{\mathrm{loc}}$ for both reflection and rotation score maps. The classication loss $\mathcal{L}_{\mathrm{cls}}$ for the intermediate predictions is the cross-entropy loss. 
The final objective $\mathcal{L}$ are expressed as
\begin{align}
    \mathcal{L}_{\mathrm{loc}} &= 
    \mathcal{L}_{\mathrm{focal}}({\mathbf{Y}}, \mathbf{Y}_\mathrm{gt}) \\
    \mathcal{L}_{\mathrm{cls}} &= 
    \mathcal{L}_{\mathrm{ce}}({\mathbf{S}}, \mathbf{S}_\mathrm{gt}) \\
    \mathcal{L} &= \mathcal{L}_{\mathrm{loc}} 
                    + \mathcal{L}_{\mathrm{cls}},
\end{align}
The network for reflection and rotation symmetry detection are denoted as {\Model}-$\mathrm{ref}$ and {\Model}-$\mathrm{rot}$, respectively.
To alleviate the class imbalance issue, the loss of the background class is weighted with $w$ for $\mathcal{L}_{\mathrm{cls}}$. Note that the focal loss alleviates the class imbalance issue for $\mathcal{L}_{\mathrm{loc}}$.


\section{New Symmetry Dataset (DENDI)}
\label{sec:dataset}


\begin{table}[t!]
\centering
\resizebox{0.49\textwidth}{!}{%
\begin{tabular}{l|ccc}
\toprule
dataset & Ref. split & Rot. split &  complexity \\ 
\midrule
SDRW~\cite{liu2013symmetry} & 51 / - / 70 & 10 / - / 66 & low \\
NYU~\cite{ConvSymm2016} & 239 / - / - & - & low \\
SymCOCO~\cite{funk20172017} & 250 / - / -& 250 / - / - & high \\
DSW~\cite{funk20172017} & - & 200 / - / - & low \\
BPS$^*$~\cite{funk2017beyond} & 959 / - / 240 & 846 / - / 211 & high \\
LDRS~\cite{seoshim2021pmcnet} & 1,110 / 127 / 240 & - & high \\
\midrule
{\Dataset} & 1,750 / 374 / 369 & 1,459 / 313 / 307 & high \\
\bottomrule
\end{tabular}
}
\vspace{-3mm}
\begin{description}
      \footnotesize
      \centering
      \item * Not available online.
\end{description}
\vspace{-5mm}
\caption{Comparison of the symmetry detection datasets.}
\vspace{-2mm}
\label{tab:datasets}
\end{table}

We present a new dataset for symmetry detection named \textit{DENse and DIverse symmetry dataset} ({\Dataset}) in the following.

\subsection{Motivation}
\paragraph{Limitations in existing datasets.}
\label{sec:ex-dataset}
The early reflection symmetry datasets~\cite{ConvSymm2016, liu2013symmetry} 
contain small number of images with few reflection axis and rotation center.
Recently proposed BPS~\cite{funk2017beyond} and LDRS~\cite{seoshim2021pmcnet} are large enough to train deep architectures, but the reflection axes still lack of diversity in terms of the length and orientation.
For example, objects with multiple symmetry axes are often annotated only by a single dominant axis.
Furthermore, no existing reflection symmetry datasets take into account the continuous symmetry group, such as a circle with an infinite number of reflection symmetry axes.
For the rotation symmetry, the annotations of BPS~\cite{funk2017beyond} 
are limited to the rotation centers while the pioneer unsupervised methods~\cite{lee2008rotation, lee2009skewed} tackle the rotation folds also. 

\paragraph{Proposed dataset.}
To address the concerns mentioned above, we present a new dataset for reflection and rotation symmetry detection that includes a wide range of geometries. 
We integrate 239 images of NYU~\cite{ConvSymm2016} and 181 images of SDRW~\cite{liu2013symmetry}, and collect 2,080 images from COCO~\cite{lin2014microsoft} dataset.
Both reflection and rotation annotations are labeled for each image, and we remove images without any labels.
As a result, {\Dataset} contains 2,493 and 2,079 images for reflection and rotation split, respectively.
The sizes of the symmetry detection datasets are compared in ~\Tbl{datasets}.
To add reflection axes with diverse length and orientation, we annotate objects in common shapes such as circle, ellipse, and polygons, as well as the part-level symmetries. 
Also, the annotators are encouraged to exhaustively label the symmetries for each object, including the non-dominant ones, {\eg} the diagonals of a square. 
For the reflection symmetry of a continuous symmetry group, we annotate with ellipse-shaped masks to represent an infinite number of line axes.
For the rotation symmetry, we additionally collect the number of rotation folds for each rotation center.
As a result, the annotations in {\Dataset} are denser and more diverse compared to the ones in the existing datasets.

\begin{figure}[t!]
    \centering
    \includegraphics[width=0.48\textwidth]{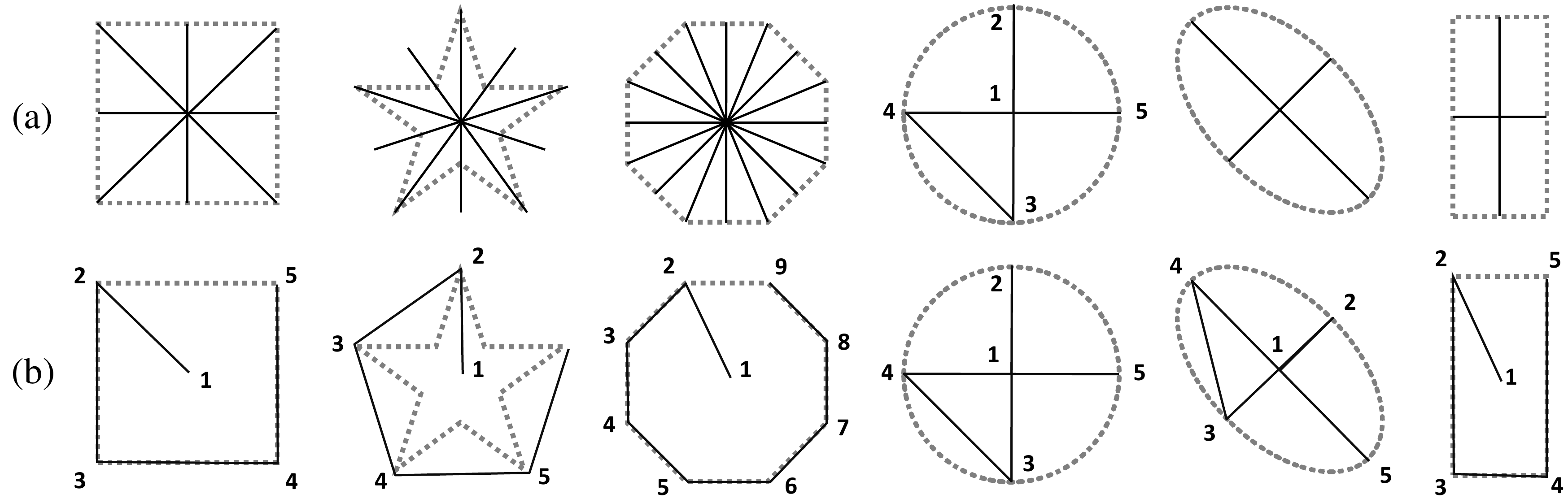}
    \caption{
    Illustration of the generic shapes and their annotations.
    (a) and (b) indicate the annotation rules of the reflection and rotation symmetry, respectively.
    For details, see ~\Sec{annotation} 
    }
    \vspace{-2mm}
    \label{fig:ann_ex}
\end{figure}

\begin{figure}[t!]
    \centering
    \includegraphics[width=0.48\textwidth]{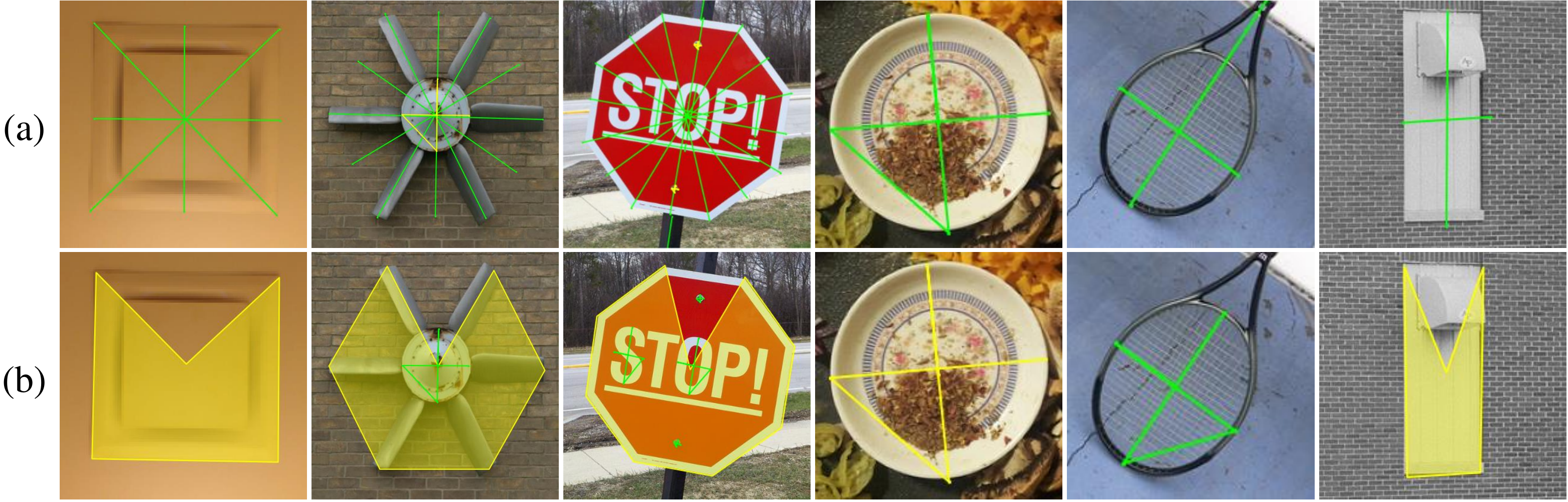}
    \caption{
    The images and labels of the objects with generic shapes.
    (a) and (b) indicate the annotations of the reflection and rotation symmetry, respectively.
    Best viewed in the electronic version.
    }
    \vspace{-2mm}
    \label{fig:ann_real_ex}
\end{figure}


\begin{figure*}[t]
    \centering
    \includegraphics[width=0.99\textwidth]{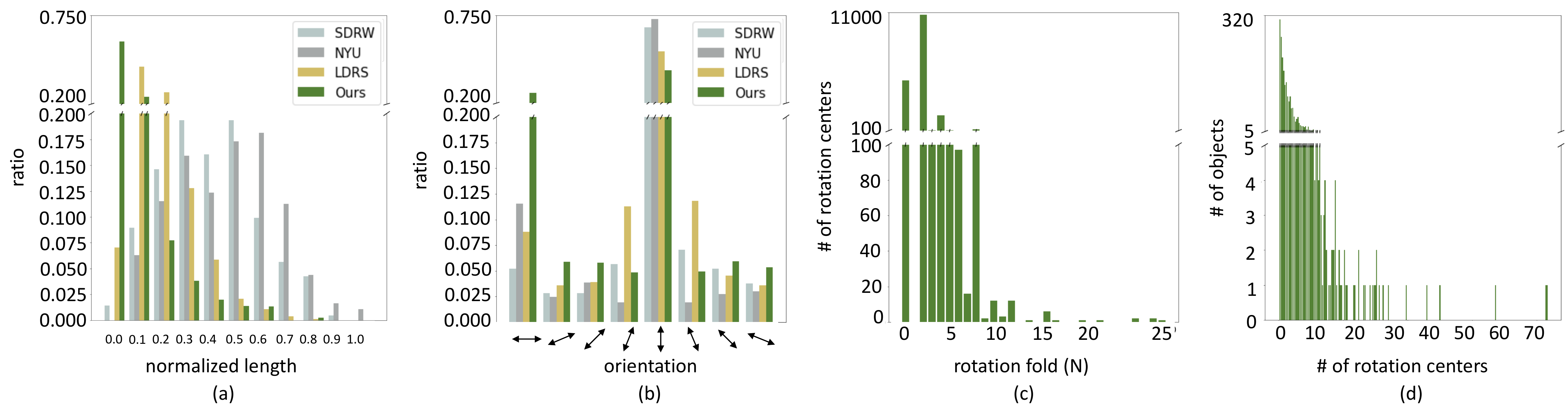}
    \vspace{-2mm}
    \caption{Statistical analysis of {\Dataset}. (a) and (b) represent the reflection symmetry dataset while (c) and (d) indicate the rotation symmetry dataset.
    In specific, (a) and (b) are histograms for scale and orientation of the reflection axes, (c) and (d) represent the histograms of the rotation fold and the number of the rotation centers.}\label{fig:chart}
    \vspace{-2mm}
\end{figure*}

\subsection{Annotation}
\label{sec:annotation}
\paragraph{Reflection symmetry.}
A reflection symmetry axis is defined as a line formed by two points following~\cite{funk20172017, funk2017beyond, seoshim2021pmcnet, ConvSymm2016, liu2013symmetry}. In contrast to the existing datasets, we now account for circular objects. 
A circular object, which is equivalent to a filled circle, has an infinite number of 
reflection symmetry axes through its center.
We propose to annotate the circular objects with 5 connected points, resembling the shape of the Arabic number '4'.
We draw a '4'-shaped annotation from the circle's center towards the circle's boundary in the up, down, left, and right directions.
The annotation rules and visualizations of the reflection symmetry for generic shapes are visualized in ~\Fig{ann_ex} (a) and ~\Fig{ann_real_ex} (a), respectively. 

\paragraph{Rotation symmetry.}
We collect the rotation center coordinate, the object's boundary, and the number of folds (N) for each object.
A circular object is in a continuous rotation group with infinite folds. Thus, we set the N as 0 for simplicity.
We categorize the objects into ellipses and polygons based on the their shape.
Circular or elliptical objects are marked with '4'-shaped annotation.
We annotate the polygons of V vertices with (V+1) consecutive points.
From the center of the object, we take the vertex closest to 12 o'clock as the 2nd point and link the vertices of a convex polygon counter clockwise.
Note that the number of the vertices (V) and the number of the folds (N) not always match. 
The annotation rules and visualizations of the rotation symmetry for generic shapes are visualized in ~\Fig{ann_ex} (b) and ~\Fig{ann_real_ex} (b), respectively. 

\subsection{Statistics}
\paragraph{Reflection symmetry.}
Histograms for scale and orientation of the reflection axes are presented in ~\Fig{chart} (a) and (b).
For scale, we measure the length of each line and normalize it with the length of the image diagonal. 
With two points on the line annotation, we can also compute its orientation (tangents). 
The y-axis of ~\Fig{chart} (a) and (b) is the ratio of the number of the axes over the total number of the axes.
The number of axes increases when the length of each axis decreases, reflecting the characteristics of the real-world environment. 
In addition, it can be interpreted that part-level symmetry is densely annotated compared to other datasets.
Note that our dataset ranks first at all orientations except for the three orientations that are closest to the vertical direction.
Despite the fact that LDRS~\cite{seoshim2021pmcnet} is also based on COCO~\cite{lin2014microsoft}, the distribution of the orientation of the axis is more diverse in our case due to the fact that we particularly request the annotators not to omit the non-dominant axes.

\paragraph{Rotation symmetry.}
Histograms of the rotation fold and the number of the rotation centers are illustrated in ~\Fig{chart} (c) and (d).
The three most common folds in the dataset are 2, 0 (continuous), and 4.
The result is predictable as there are many rectangles, circles, and squares in the image.
Note that the dataset contains a notable number of objects of fold 8, which are mostly the 'STOP' signs in the road.
The complexity of our rotation symmetry dataset is high, as shown in ~\Fig{chart} (d).
Even if we clip a few exceptions, a lot of rotation centers are marked in the images.


\section{Experiments}
\label{sec:experiments}

\subsection{Experimental settings}
\paragraph{Datasets.}
We use SDRW~\cite{liu2013symmetry}, LDRS~\cite{seoshim2021pmcnet}, and {\Dataset} to evaluate the reflection symmetry detection model.
We follow the training and evaluation settings of PMCNet~\cite{seoshim2021pmcnet} in ~\Tbl{ldrs_compare} for additional use of NYU~\cite{ConvSymm2016} and the synthesized images. 
For rotation, we only use {\Dataset} which contains the images of SDRW~\cite{liu2013symmetry} rotation dataset.

\paragraph{Evaluation.}
\label{sec:eval}
To evaluate {\Model}, we use F1-score computed with the precision and recall as 
$\frac{2 \times \mathrm{prec} \times \mathrm{rec}}{\mathrm{prec} + \mathrm{rec}}$. 
A convention~\cite{funk2017beyond, seoshim2021pmcnet, tsogkas2012learning} for measuring the score of the output score map is morphological thinning~\cite{martin2004learning} and an off-the-shelf pixel-matching algorithm to compare with the ground-truth lines, which are also pixel-width.
In contrast to the existing datasets, we take the circular objects into account, which are annotated with filled circles. 
As {\Dataset} contains annotations of filled circles, the thinning operation shrinks to a single-pixel dot. 
Therefore, it is inevitable to come up with a new way to evaluate the predicted score maps. 
We dilate the ground-truth score maps of reflection and rotation symmetries with a maximum distance of 5 pixels following~\cite{funk2017beyond} where they enlarge the ground-truth dots to 5-pixel radius circles around the rotation centers.
The predicted score map is also dilated to make the result of the ground-truth itself become 1. 
The true-positives are then computed by pixel-wise comparisons.

\paragraph{Implementation details.}
As a backbone network, we adopt the ReResNet implementation of ReDet~\cite{han2021redet} of depth 50, which is based on PyTorch~\cite{pytorch} and e2cnn~\cite{e2cnn}. 
The number of the layers and the structure are the same as the vanilla ResNet~\cite{he2016deep}.
We pretrain the ReResNet50 for the image classification task in imagenet-1k~\cite{deng2009imagenet} following the procedures in ~\cite{han2021redet}.
For the symmetry group to initialize the equivariant networks, we use a dihedral group of eight orientations ($\mathrm{D}_8$).
To provide multi-scale contexts, we also deploy the Atrous Spatial Pyramid Pooling module (ASPP)~\cite{DBLP:journals/corr/ChenPSA17} which we re-implemented the module by replacing all the vanilla convolutions to E(2)-equivariant convolutions~\cite{e2cnn}. 
The number of the classes $N^\mathrm{ref}$ is 8 and $N^\mathrm{rot}$ is 21.
We train {\Model} for 100 epochs with an initial learning rate of 0.001 using the Adam~\cite{Adamsolver} optimizer with a batch size of 32.
For details, refer to the supplementary material.


\begin{table}[t!]
\centering
\resizebox{0.31\textwidth}{!}{%
\begin{tabular}{l||cc|cc}
\toprule
\multirow{2}{*}{model} & \multicolumn{2}{c|}{design choices}   & \multicolumn{2}{c}{F1 score} \\
& Equiv. & Aux. &  Ref. & Rot. \\ 
\midrule
\multirow{3}{*}{Ref.} & &  & 55.1 & - \\
 & \checkmark &  & 63.1 & - \\
 & \checkmark &  \checkmark & \textbf{64.5}  & - \\
\midrule 
\multirow{3}{*}{Rot.} &  &   & -  & 17.7 \\
 & \checkmark &   &   -  & 21.2 \\
 & \checkmark & \checkmark &  -  &  \textbf{22.5} \\  
\midrule
\multirow{2}{*}{joint}  & \checkmark &   & 62.2 & 22.1 \\
 & \checkmark & \checkmark  & 58.7 & 22.5 \\  
\bottomrule
\end{tabular}
}
\caption{Ablation on the symmetry detection network in reflection, rotation, and joint model on {\Dataset} dataset.}
\vspace{-4mm}
\label{tab:ab-all}
\end{table}

\subsection{Ablation studies}
\paragraph{Ablation on the equivariant convolution.} 
We study the effectiveness of the group-equivariant convolution in ~\Tbl{ab-all}.
With the help of the equivariant convolutions the F1 scores of 55.1 and 17.7 increase to 63.1 and 21.2 for {\Model}-$\mathrm{ref}$ and {\Model}-$\mathrm{rot}$, respectively. 

\paragraph{Ablation on the auxiliary classification.}
To enhance the intermediate representation, we perform a relevant subtask for each branch of symmetry detection and compare them in ~\Tbl{ab-all}.
Without extra labels, the reflection-only model achieves the F1 score of 64.5, which is greater than the 63.1 obtained by training solely with the final task.
The orientation estimation also requires rotation equivariance, which enhances the intermediate features.
Rotation symmetry detection, on the other hand, requires extra annotation for the auxiliary task as the original labels are a set of dots. 
The rotation invariant subtask of classification of the rotation folds (N) compresses the information of the intermediate feature so that it can increase the F1 score from 21.2 to 22.5.

\paragraph{Ablation on the joint training.}
We investigate the effect of a joint training of the reflection and rotation symmetries in ~\Tbl{ab-all}.
When training {\Model}-$\mathrm{joint}$, the loss $\mathcal{L}$ is 
computed for both reflection and rotation symmetries. 
Joint symmetry detection network that is trained only with the final task achieves comparable F1 scores of 62.2 and 22.1 for reflection and rotation symmetry, respectively. 
However, the auxiliary task do not increase the accuracy of the reflection symmetry in joint training scenario. 
One probable explanation is that only orientation estimation of the reflection symmetry axis requires rotation equivariance while the fold classification only necessitates rotation invariance, resulting in network imbalance.

\begin{table}[t!]
\centering
\resizebox{0.33\textwidth}{!}{%
\begin{tabular}{l||l|c}
\toprule
symmetry & method & F1 score \\
\midrule
\multirow{3}{*}{reflection} & 
 {SymResNet$^{*}$~\cite{funk2017beyond}} & 30.7 \\
 & {PMCNet$^{\dag}$~\cite{seoshim2021pmcnet}} & 52.0 \\
 & {{\Model}-$\mathrm{ref}$}  & \textbf{64.5} \\
\midrule
\multirow{2}{*}{rotation} 
 & {SymResNet$^{*}$~\cite{funk2017beyond}} & 11.9 \\
 & {{\Model}-$\mathrm{rot}$} & \textbf{22.5} \\
\bottomrule
\end{tabular}
}
\vspace{-1mm}
\begin{description}
      \footnotesize
      \centering
      \item \dag Re-trained on DENDI.
        \vspace{-2mm}
      \item {*} Evaluated using the weights provided by the authors.
\end{description}
\vspace{-5mm}
\caption{Comparison with the state-of-the-art methods on {\Dataset}.}
\label{tab:sota}
\end{table}
 
\begin{table}[t]
\centering
\small
\resizebox{0.4\textwidth}{!}{%
\setlength\tabcolsep{3pt}
\begin{tabular}{l||cc|cc||c}
\toprule
\multirow{2}{*}{method}
&  \multicolumn{2}{c|}{train dataset} &  \multicolumn{2}{c||}{test dataset} &
\multirow{2}{*}{$\mathrm{mixed}$} \\ 
 & $\mathrm{real}$ & $\mathrm{synth}$ & SDRW & LDRS &  \\ 
\midrule
\multirow{2}{*}{PMCNet~\cite{seoshim2021pmcnet}}  & \checkmark &  & 61.6 & 34.8 & \multirow{2}{*}{61.2} \\
  & \checkmark & \checkmark & \textbf{68.8} & 37.3 & \\
\midrule
\multirow{2}{*}{{\Model}-$\mathrm{ref}$}  & \checkmark &  & 67.4 & \textbf{40.9} & \multirow{2}{*}{\textbf{71.4}}\\
 & \checkmark & \checkmark & 67.1 & 39.4 & \\
\bottomrule
\end{tabular}%
}
\vspace{-2mm}
\caption{Comparison of the reflection symmetry detection methods on the LDRS~\cite{seoshim2021pmcnet} and SDRW~\cite{liu2013symmetry}. Note that the $\mathrm{real}$ dataset consists of SDRW, LDRS, and NYU~\cite{ConvSymm2016} dataset.}
\vspace{-2mm}
\label{tab:ldrs_compare}
\end{table}

\begin{figure*}[t]
    \centering
    \includegraphics[width=\textwidth]{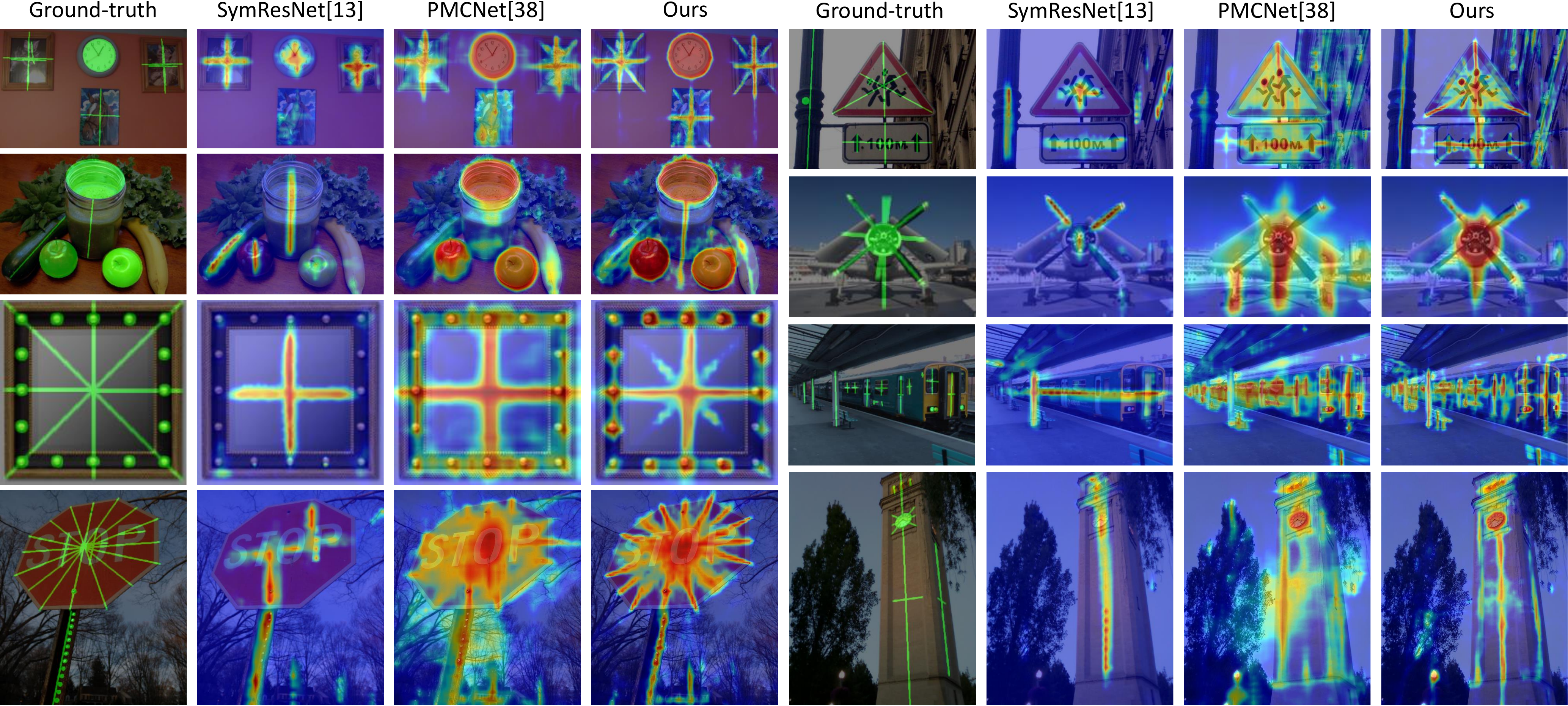}
    \caption{Qualitative results of the reflection symmetry detection on {\Dataset}-$\mathrm{ref}$ \textit{test}. }
    \label{fig:qual_ref}
\end{figure*}

\begin{figure*}[t]
    \centering
    \includegraphics[width=\textwidth]{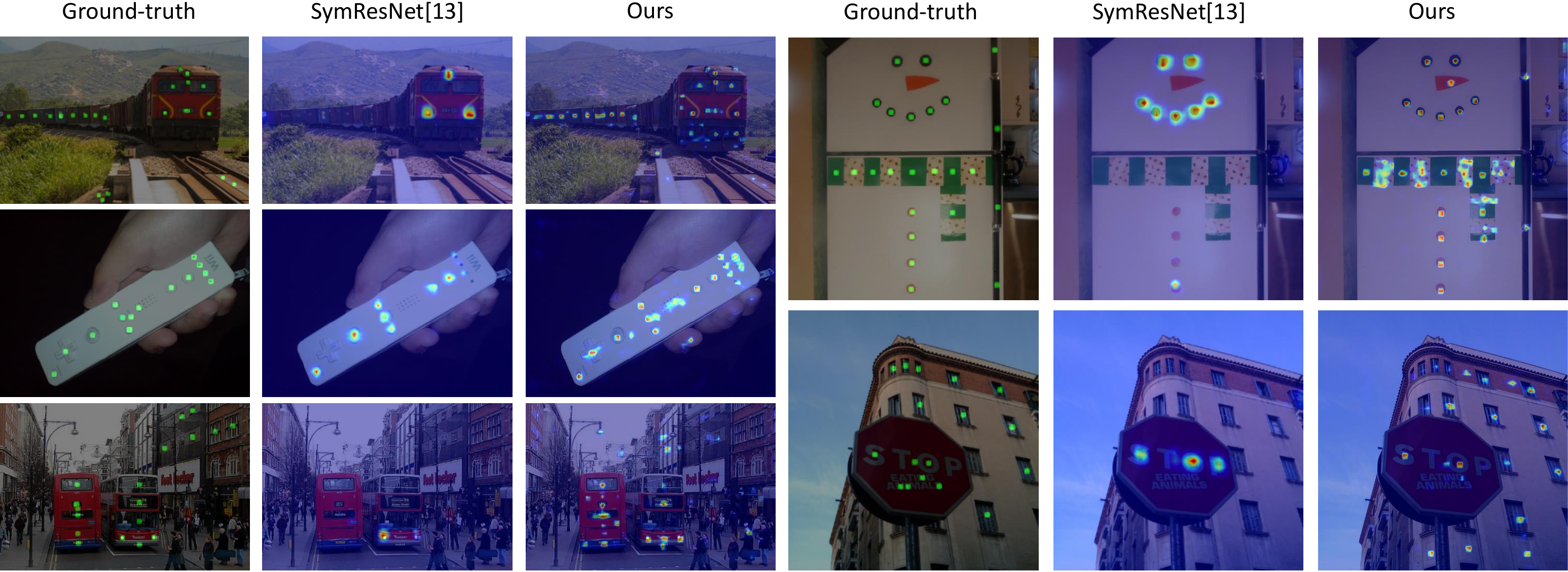}
    \caption{Qualitative results of the rotation symmetry detection on {\Dataset}-$\mathrm{rot}$ \textit{test}.}
    \label{fig:qual_rot}
\end{figure*}

\subsection{Comparison with the state-of-the-art methods}
We compare {\Model} with the state-of-the-art methods in~\Tbl{sota} and~\Tbl{ldrs_compare}.
For both reflection and rotation symmetries, our proposed {\Model} achieves the state-of-the-art, showing the effectiveness of the equivariant networks and the auxiliary classification.
While PMCNet~\cite{seoshim2021pmcnet} is re-trained on {\Dataset} for a fair comparison, 
SymResNet~\cite{funk2017beyond} is compared using the weights provided by the authors as
fine-tuning SymResNet~\cite{funk2017beyond} degraded the performance.

We follow the configurations of PMCNet~\cite{seoshim2021pmcnet} for the experiments on SDRW~\cite{liu2013symmetry} and LDRS~\cite{seoshim2021pmcnet} in~\Tbl{ldrs_compare}
The training images consist of $\mathrm{real}$ images from SDRW, LDRS, and NYU~\cite{ConvSymm2016} and the generate $\mathrm{synthetic}$ images as in~\cite{seoshim2021pmcnet}. 
{\Model}-$\mathrm{ref}$ achieves the state-of-the-art on LDRS while the results on SDRW are still comparable. 
The additional use of synthetic images is not helpful as proposed in~\cite{seoshim2021pmcnet}. 
One possible reason is the imbalance in data distribution across splits.
To mitigate this issue, we construct a new split denoted as $\mathrm{mixed}$ by merging all images and then randomly split the images into \textit{train/val/test} splits with the ratio of 4:1:1.
{\Model}-$\mathrm{ref}$ outperforms PMCNet in that scenario, as shown in ~\Tbl{ldrs_compare}.
All the experiments in~\Tbl{ldrs_compare} are evaluated with the legacy scheme.
 
\subsection{Qualitative results}
The qualitative results of {\Model}-$\mathrm{ref}$, PMCNet~\cite{seoshim2021pmcnet}, and SymResNet~\cite{funk2017beyond} on {\Dataset}-$\mathrm{ref}$ \textit{test} are shown in~\Fig{qual_ref}. 
{\Model}-$\mathrm{ref}$ produces dense reflection symmetry score maps compared to other methods, including non-dominant axes such as the diagonal. 
Furthermore, {\Model}-$\mathrm{ref}$ predicts masks of the circular objects accurately
even for the challenging samples where the line and circles both exist in the ground-truth.
We compare {\Model}-$\mathrm{rot}$ with SymResNet on {\Dataset}-$\mathrm{rot}$ \textit{test} in~\Fig{qual_rot}. 
{\Model}-$\mathrm{rot}$ is robust in scale and number of rotation centers.
{\Model}-$\mathrm{rot}$ detects symmetry of polygons as well as circular objects in {\Dataset}-$\mathrm{rot}$, whereas SymResNet mainly detects circular objects.

\subsection{Limitations}
While {\Model} can be jointly trained to produce comparable predictions as the single-branch {\Model}, it has more room for improvement.
Especially, the design of the auxiliary task of {\Model}-$\mathrm{rot}$ can be explored more to enhance the accuracy of the reflection symmetry detection.


\section{Conclusion}
In this paper, we have proposed a novel symmetry detection framework, {\Model}, using equivariant learning to obtain group-equivariant and invariant scores for both reflection and rotation symmetries.
In addition, we have introduced a new dataset \textit{DENse and DIverse symmetry dataset} ({\Dataset}) for reflection and rotation symmetries. 
The proposed {\Model} achieves the state-of-the-art on LDRS and {\Dataset} datasets.

\smallbreak
\noindent \textbf{Acknowledgements.}
This work was supported by Samsung Advanced Institute of Technology (SAIT) and also by the NRF grant (NRF-2021R1A2C3012728) and the IITP grant (No.2021-0-02068: AI Innovation Hub, No.2019-0-01906: Artificial Intelligence Graduate School Program at POSTECH) funded by the Korea government (MSIT).  
We like to thank Yunseon Choi for her contribution to {\Dataset}.
\looseness=-1

{\small
\bibliographystyle{ieee_fullname}
\bibliography{main}
}

\clearpage
\appendix

\setcounter{section}{0}
\setcounter{figure}{0}
\setcounter{table}{0}

\renewcommand{\thesection}{A.\arabic{section}}
\renewcommand{\thefigure}{a.\arabic{figure}}
\renewcommand{\thetable}{a.\arabic{table}}

\section*{Appendix}


\section{EquiSym}

The details that are omitted in the main paper are covered in this section.
We show the consistency of the evaluation schemes. 
The implementation details are also shown in the following.

\subsection{Evaluation scheme}
\begin{table}[h!]
\centering
\small
\resizebox{0.42\textwidth}{!}{%
\setlength\tabcolsep{3pt}
\begin{tabular}{l||cc|cc||c}
\toprule
\multirow{2}{*}{method}
&  \multicolumn{2}{c|}{train dataset} &  \multicolumn{2}{c||}{test dataset} &
\multirow{2}{*}{$\mathrm{mixed}$} \\ 
 & $\mathrm{real}$ & $\mathrm{synth}$ & SDRW & LDRS &  \\ 
\midrule
\multirow{2}{*}{PMCNet~\cite{seoshim2021pmcnet}}  & \checkmark &  & N/A & N/A & \multirow{2}{*}{46.6} \\
  & \checkmark & \checkmark & \textbf{51.1} & 33.7 & \\
\midrule
\multirow{2}{*}{{\Model}-$\mathrm{ref}$}  & \checkmark &  & 49.8 & \textbf{36.5} & \multirow{2}{*}{\textbf{58.0}}\\
 & \checkmark & \checkmark & 49.2 & 34.9 & \\
\bottomrule
\end{tabular}%
}
\vspace{-2mm}
\caption{Comparison of the reflection symmetry detection methods on the LDRS~\cite{seoshim2021pmcnet} and SDRW~\cite{liu2013symmetry}. Note that the $\mathrm{real}$ dataset consists of SDRW, LDRS, and NYU~\cite{ConvSymm2016} dataset and $\mathrm{synthetic}$ images generated as in~\cite{seoshim2021pmcnet}. }
\vspace{-2mm}
\label{tab:ldrs_compare}
\end{table}

In the main paper, we propose to use a modified evaluation scheme of blurring the ground truth rather than thinning the prediction. The primary reason is that the thinning process transforms a circular mask prediction into a single dot. The pixel matching algorithm determines whether the predicted lines are close enough to the ground truth lines within a threshold, which becomes equivalent when the ground truth itself is blurred with a radius of a threshold. 
In practice, we construct a filter of the kernel $11 \times 11$ where the weights are set to 1 for a circle of a diameter $11$ and 0 otherwise. 
We convolve the ground-truth heatmap of both symmetries with the filter so the heatmap is dilated to the maximum distance of 5 pixels.
The true positives are then computed by pixel-wise comparisons. 
We re-evaluate the experiments of Tab. 3 of the main paper in ~\Tbl{ldrs_compare}. Note that one experiment from PMCNet~\cite{seoshim2021pmcnet} is excluded since the trained model is not accessible.
As shown in~\Tbl{ldrs_compare}, the rankings produced by the two evaluation schemes are consistent, while the latter is significantly faster.

\subsection{Implementation details}

\paragraph{Construction of the orientation labels.}
{\Model} utilizes the intermediate tasks to increase the accuracy of the symmetry detection tasks. In the case of reflection symmetry, the intermediate labels of the orientations of the reflection axes are obtained for free. 
The angle of the reflection symmetry axis in the form of a straight line can be expressed as a linear combination of the closest one or two angles among the 8 predetermined angles, which is an initial soft label.
On the other hand, the circle-shaped symmetry axis has an evenly divided orientation label of 8 segments determined by the orientation of the line crossing the center.
The orientation label is then quantized for training.

\paragraph{ImageNet pretrain.}
To be consistent with experiments on the vanilla ResNet~\cite{he2016deep} pre-trained on ImageNet~\cite{deng2009imagenet}, we pretrain the ReResNet50 on ImageNet-1K for the image classification.
While ReResNet50 from ReDet is implemented with $\mathrm{C}_8$ group, we use $\mathrm{D}_8$ group instead.
Furthermore, we adjust the stride and dilation of each layer in ReResNet50 to obtain a higher resolution feature map with a larger receptive field than the original one, which is a common procedure in the semantic segmentation~\cite{DBLP:journals/corr/ChenPSA17}.
The learning rate starts at 0.1 and decreases by 0.1 every 30 epochs, for a total of 100 epochs. We use a batch size of 512.
The pretrained ReResNet50 achieves 69.06\% top-1 and 87.25\% top-5 accuracy on the ImageNet \textit{val}.

\paragraph{Implementation details.}
Following ~\cite{seoshim2021pmcnet}, The hyperparameters $\alpha$ and $\beta$ of the focal loss are set to 0.95 and 2, respectively.
For training, we resize input images so that the maximum length of the width or the height is 417. 
The background weight $w$ of $\mathcal{L}_{\mathrm{cls}}$ is set to 0.01 and 0.001 for reflection and rotation symmetries.
We use the PyTorch~\cite{pytorch} and e2cnn~\cite{e2cnn} framework to build our model. 


\section{DENDI}
\begin{figure*}[t!]
    \centering
    \scalebox{0.93}{
    \centering
    \includegraphics[width=\textwidth]{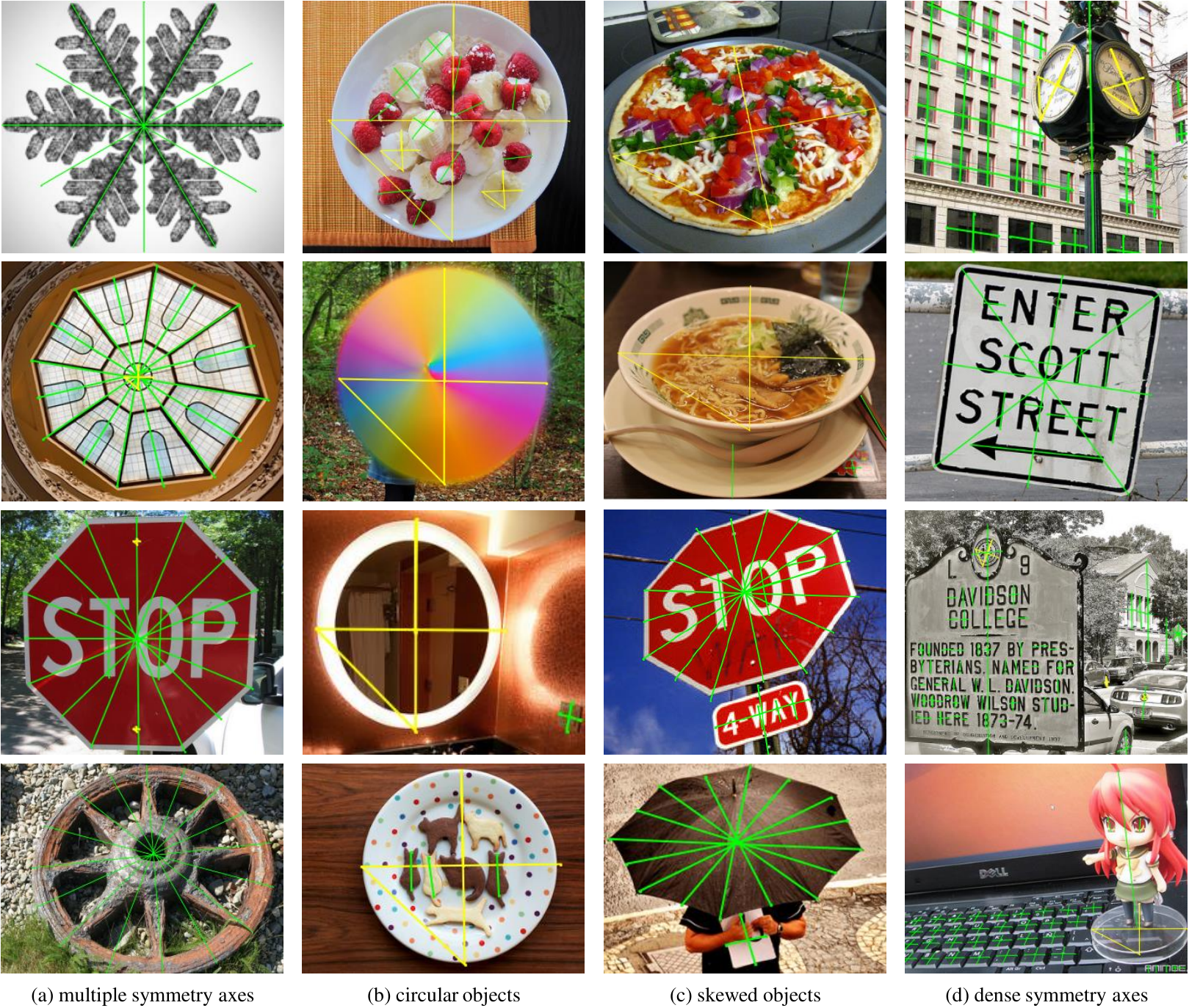}
    }
    \caption{Illustration of the examples in the reflection symmetry dataset. 
    The samples with (a) multiple symmetry axes, (b) circular objects, (c) skewed objects, and (d) dense symmetry axes are shown in the figure. Green lines indicate the reflection axes and the yellow lines indicate the '4'-shaped reflection circle annotation. The reflection-circle annotations are then converted to masks.
    }
    \label{fig:ann_ex_ref}
\end{figure*}

\begin{figure*}[t!]
    \centering
    \scalebox{0.93}{
    \centering
    \includegraphics[width=\textwidth]{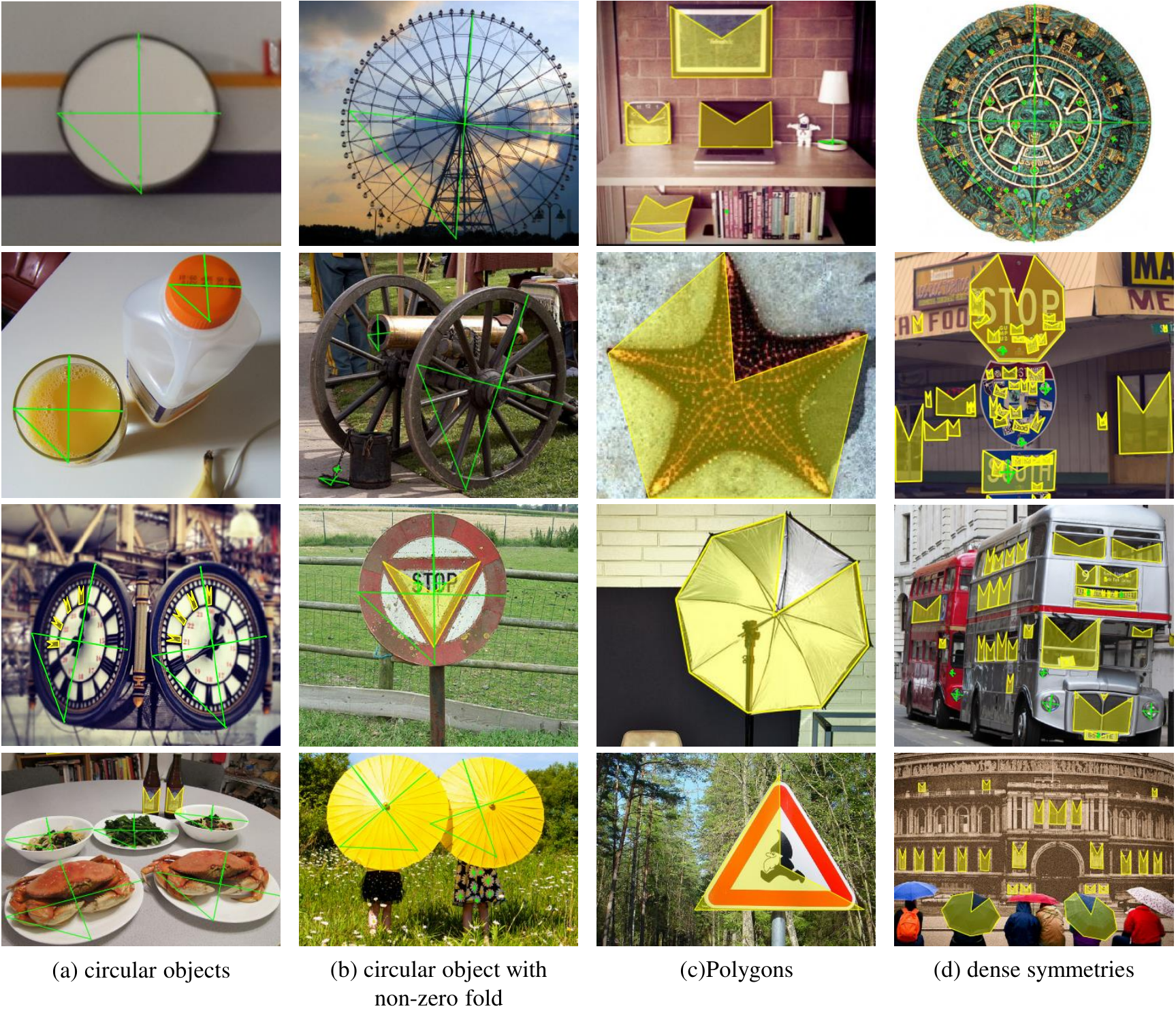}
    }
    \caption{Illustration of the examples in the rotation symmetry dataset. The samples with  (a) circular objects, (b) circular objects with folds larger than 2, (c) polygons, and (d) dense symmetries are shown in the figure. Green lines indicate the circular annotations and the yellow polygons indicate the polygon-type annotations. Only the center coordinates are used for evaluation.
    }
    \label{fig:ann_ex_rot}
\end{figure*}

The details of data annotation for the DENDI dataset are described in this section. 
To identify symmetry, we disregard texture and focus just on the shape of the object. 
The partial occlusion of boundary of symmetric object is allowed to a fourth of the object boundaries. 
For both reflection and rotation symmetries, we exhaustively mark all symmetries in an image, including those of parts. This policy ensures that the DENDI contains dense annotations. we present some examples in ~\Fig{ann_ex_ref} (d) and ~\Fig{ann_ex_rot} (d). 

\subsection{Reflection symmetry}
A reflection symmetry axis is drawn as a line following the previous datasets~\cite{funk20172017, funk2017beyond, seoshim2021pmcnet, ConvSymm2016, liu2013symmetry}. The notable examples are shown in~\Fig{ann_ex_ref} (a). 
We annotate all reflection symmetry axes in an object, including non-dominant ones. Different from the existing datasets, we account for a circular object, which has an infinite number of reflection symmetry axes. 
Instead of an infinite number of symmetry axes to represent a circular object, we use a '4'-shaped label consisting of 5 points which are then converted to a circular mask, as shown in~\Fig{ann_ex_ref} (b).
Note that a semantically circular object that seems to be an ellipse due to viewpoint variants is also annotated with a '4'-shaped label, as show in the first tow rows in~\Fig{ann_ex_ref} (c).
Likewise, a skewed regular polygon due to perspective variations has the same reflection axis as a non-skewed regular polygon, as shown in the last two rows in~\Fig{ann_ex_ref} (c),~\eg, a regular STOP sign and a skewed STOP sign that are both semantically regular octagon have eight reflection symmetry axes.

Furthermore, we annotate symmetry in characters such as A, B, C, D, E, H, I, K, M, O, T, U, V, W, X, and Y, as well as the numbers 0, 1, 2, 5, and 8, except for those that are too thin or indistinct. We also annotate symmetry in the D-shaped part of characters, such as P and R.
If multiple symmetry axes are overlapped, only the longest one is saved.

\subsection{Rotation symmetry}
For each object with rotation symmetry, we collect the coordinate of the rotation center, the boundary of the object, and the number of the rotation folds (N).
We again employ the '4'-shaped labels to denote circular or elliptical objects as shown in~\Fig{ann_ex_rot} (a). 
The semantically circular object also features an infinite number of rotation folds, indicated as 0 for simplicity, in addition to the '4'-shaped labels. Similar to reflection symmetry, an semantically circular object with elliptical shape due to viewpoint variants has a rotation fold of 2,~\eg, the third and fourth rows in~\Fig{ann_ex_rot} (a). 
The minor axis takes precedence over the major axis when drawing '4'-shaped labels for elliptical objects.
The rotation fold of a circular object, in particular, can be greater than 2 if the object contains cyclic symmetry.
In the case of a non-circular object with rotation symmetry such as~\Fig{ann_ex_rot} (c), we draw a convex polygon starting from the center of the object and following convex vertices. The vertex nearest to 12 o'clock takes priority among the convex vertices.
Likewise, in the reflection symmetry dataset, symmetry in characters such as H, I, N, O, S, X, and Z, as well as the numbers 0 and 8, are taken into account.

\end{document}